# Efficient Training of Robust Traditional Chinese LLaMA-1B on a Single Consumer GPU: Continual Pre-training, SFT, and DPO

Yu-Cheng Chih[1]*

*International Intercollegiate Ph.D. Program. National Tsing Hua University, Taiwan*
*NLP Algorithm Engineer*
*Former Engineer, Accton Technology Corporation*
Email：s111003816@m111.nthu.edu.tw

Ming-Tao Duan[2]

*EECS International Graduate Program. National Yang Ming Chiao Tung University, Taiwan*
Email：alvisaung@gmail.com

Yong-Hao Hou[2]

*Department of Computer Science, University of Taipei, Taiwan*
*AI Algorithm Engineer*
Email：g11016011@go.utaipei.edu.tw

## Abstract

Small Language Models (SLMs) enable cost-effective, on-device and latency-sensitive AI applications, yet their deployment in Traditional Chinese (TC) remains hindered by token-level instability—models unpredictably emit non-TC characters or code-switch into other languages. We address this practical reliability gap by creating **PureTC-1B**, a three-stage stabilization pipeline for **Llama-3.2-1B-Instruct** (an open-weight, instruction-tuned model released by Meta) **[1]** using parameter-efficient LoRA adapters [2]. Our method combines **Continual Pre-Training (CPT)** on TC-centric corpora, **Supervised Fine-Tuning (SFT)** with instruction data, and **Direct Preference Optimization (DPO)** [3] using TC-adherence preferences to improve monolingual robustness without full-model retraining.

On a benchmark designed to simulate real-world usage, PureTC-1B achieves a **51.3% relative reduction** (micro-average) in non-TC output tokens versus the base model. On a Named Entity Translation (NET) task, PureTC-1B **further reduces incorrect-language tokens by 77.2% relative to Llama-3B** and 57.2% relative to Qwen-1.5B, indicating that robust

TC adherence is attainable even at the 1B scale. The pipeline is reproducible, adapter-only, and hardware-friendly, offering practitioners a practical recipe to enhance language stability for TC and potentially other non-English languages.

## 1. Introduction

The rapid progress of open-source Large Language Models (LLMs), notably the Llama family, has catalyzed a shift toward domain-specialized AI systems. Within this trend, Small Language Models (SLMs) offer a compelling trade-off between accuracy, cost, and deployability, enabling use cases in public administration, legal services, and education. However, effective deployment in Traditional Chinese (TC) settings hinges on a property that is easy to desire but hard to guarantee: token-level linguistic stability. In practice, many open-weight models—across sizes from ~1B to ~7B—exhibit spurious code-switching or emit non-TC characters even under TC prompts, undermining reliability for production [4].

A We target this overlooked yet deployment-critical failure mode. We introduce PureTC-1B, an adapter-only stabilization pipeline for Llama-3.2-1B-Instruct [1] that enforces TC adherence without expensive full-model retraining. The pipeline proceeds in three stages: CPT augments subword statistics and character priors with TC-centric text; SFT aligns task behavior with TC instructions; DPO directly optimizes preferences for TC-consistent outputs under mixed or multilingual inputs [3]. This design emphasizes practicality (fits commodity GPUs), reproducibility (single-seed, fixed decoding), and transferability to other languages suffering from similar instability.

## 2. Problems & Contributions

*2.1. Problem Statement*

While Small Language Models (SLMs) are increasingly favored for cost-efficient deployment, their use in **Traditional Chinese (TC)** contexts is critically hindered by **token-level instability**. This failure mode manifests as unintended **code-switching**, where models insert Simplified Chinese, Japanese, or English tokens into otherwise TC outputs [4]. Such contamination significantly compromises the reliability required for professional applications in government, law, and education.

As demonstrated in our baseline experiments (§6), the open-weight Llama-3.2-1B-Instruct model frequently exhibits this instability. Under TC prompts, its responses contain a non-trivial fraction of non-TC tokens, confirming that even state-of-the-art open-source SLMs lack the linguistic stability necessary for deployment [4].

*2.2. Contributions*

Our primary contribution is a parameter-efficient, adapter-based pipeline that systematically stabilizes Traditional Chinese generation in SLMs through Continual Pre-training (CPT), Supervised Fine-Tuning (SFT), and Direct Preference Optimization (DPO). Specifically, we contribute：

- **Novel Metrics for Linguistic Stability.** We introduce a new set of metrics—Other-Language Rate (OLR) and Chinese Stability Rate (CSR)—to quantify and benchmark undesired language contamination in generated outputs.
- **A Modular Stabilization Pipeline.** We design and validate a fully reproducible CPT → SFT → DPO pipeline using LoRA adapters [2]. This modular approach significantly reduces language contamination in **LLaMA-3.2-1B-Instruct** while preserving task accuracy and ensuring simplified, mergeable deployment.
- **A Stable 1B TC Model**. On our 660-prompt benchmark under fixed decoding, PureTC-1B achieves the strongest TC-stability among 1B-class models and, on stability metrics, **outperforms larger baselines** (e.g., vs. Llama-3B and Qwen-1.5B). In particular, **OLR** drops by **51.3%** (micro) / **54.1%** (macro) relative to Llama-3.2-1B-Instruct, and **Pass@TC** rises from **9.5%→29.9%** (**+20.4 pp; ≈3.1×**), with gains in 7/8 task families.

## 3. Related Work

*3.1. Landscape of LLaMA-based Models for Traditional Chinese Applications*

Across Taiwan, researchers, companies, and government bodies are adapting Meta's LLaMA models to meet local needs in each sectors. The academic-led Taiwan-LLM project has adapted LLaMA-3-70B for specialized fields such as law, medicine, and manufacturing by training it on relevant local data [5]. In the private sector, Foxconn developed FoxBrain, an LLM based on LLaMA 3.1, is Taiwan's first traditional Chinese LLM optimized for supply chain decision support, document processing and reasoning [6]. Similarly, the government-led TAIDE initiative leverages LLaMA-2 and LLaMA-3 backbones with local data, deploying them in areas such as education, regulation, and dialogue systems [7]. These projects show a clear and growing demand for applying LLMs in important, real-world scenarios. However, while these models are being customized for specific downstream tasks, their core reliability in producing clean and consistent Traditional Chinese output remains a key challenge.

*3.2 Challenge of output Instability in Multilingual Model*

While the models discussed previously are adapted for Traditional Chinese tasks, a fundamental issue persists: the output instability inherent in many open-source multilingual LLMs. This problem, often termed token contamination or unintended code-switching, manifests as the spontaneous generation of foreign-language tokens—typically English or Southeast Asia—within an otherwise Chinese response. This linguistic leakage undermines the model's reliability.

Recent research validates the severity of this issue. A 2025 study, Lost in the Mix, demonstrated that the presence of mixed-language tokens can significantly degrade a model's reasoning and comprehension capabilities. Crucially, the authors found that fine-tuning is a far more robust solution for mitigating this instability than prompt engineering [4]. This aligns with earlier findings from EMNLP 2023, which concluded that smaller, fine-tuned models consistently outperform larger, general-purpose multilingual models on language-specific tasks [8].

These findings highlight a critical insight: model scale alone does not guarantee linguistic fidelity. For high-stakes domains such as legal document processing or educational content generation, this instability is not a minor inconvenience but a critical barrier to adoption. A single incorrect token can alter legal meaning, introduce ambiguity into official documents, or confuse a learner. Therefore, achieving stable, high-fidelity Traditional Chinese output is an essential prerequisite for building truly dependable applications.

*3.3 Techniques for Model Specialization and Alignment*

Emerging techniques in LLM fine-tuning reveal that deliberate incorporation of code-switching—whether at curriculum, synthetic, or pretraining levels—can significantly enhance language alignment and cross-lingual transfer. For example, Code-Switching Curriculum Learning (CSCL) demonstrates that training a model sequentially on token-level code-switching, then sentence-level, and finally monolingual data, significantly improves cross-lingual transfer capabilities [9]. A parallel line of inquiry—synthetic code-switched fine-tuning—demonstrates that fine-tuning on controlled, artificially mixed-language datasets (e.g., derived from CommonSenseQA) elevates performance in low-resource languages while preserving high-resource accuracy [10]. Complementing these approaches, work exploring code-switching during pretraining shows that even naturally occurring code-switching embedded within training corpora contributes to multilingual transfer capabilities; further enhancements are achieved by scaling synthetic code-switching, which improves language alignment and representation across data-tier diversity [11].

These established principles directly inform our methodology. We adopt a multi-stage pipeline, such as CPT, SFT and preference optimization pipelines, for achieving stable and aligned Traditional Chinese generation in LLaMA-based models, particularly where multilingual interference and code-switching pose application-level risk.

## 4. Methodology

*4.1 System Overview*

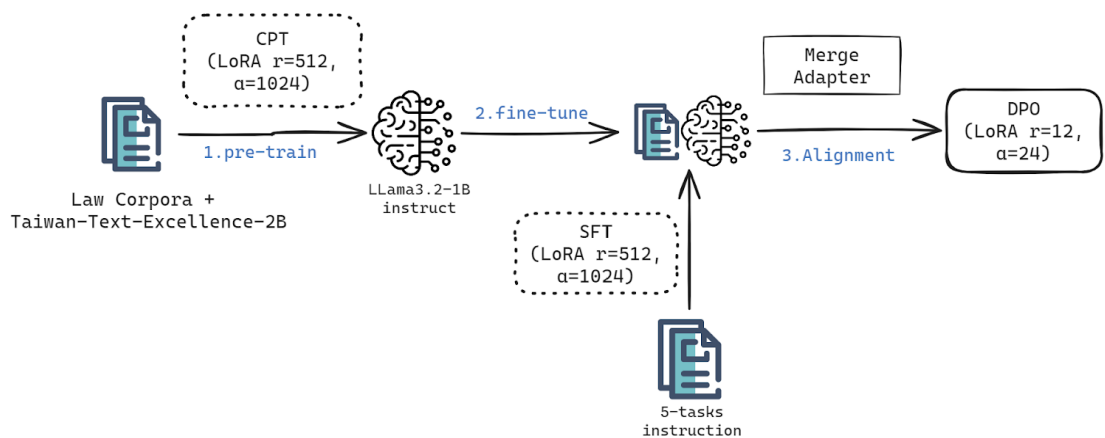

**Figure 1.** Overall Finetune Flow

Our methodology corrects unstable Traditional Chinese (TC) generation in the Llama-3.2-1B-Instruct model by applying a sequential, adapter-based fine-tuning pipeline. The entire process, illustrated in Figure 1, uses a frozen base model and consists of three stages:

- **Continual Pretraining (CPT)**: This stage shifts the model's language priors toward TC in two steps. First, an adapter is trained for one epoch on a general TC corpus to make an initial adjustment. This adapter is merged. Second, a new adapter is trained for multiple epochs on the updated base model using a specialized corpus to deepen domain alignment.

- **Supervised finetuning (SFT)**: The model is then fine-tuned on five distinct instruction tasks to improve its ability to generate contextually faithful and structured output.
- **Direct Preference Optimization (DPO) [1-3]**: Finally, a lightweight DPO stage aligns the model to prefer linguistically pure TC outputs over semantically equivalent answers containing code-switching.

Each stage utilizes LoRA for parameter-efficient adaptation, targeting all linear attention and feed-forward layers [2]. The SFT adapter is merged into the base model before a final, low-rank DPO adapter is trained. This last adapter is then merged to produce a single, deployable checkpoint.

*4.2 Base Model & Tokenizer*

Our work is based on the public meta-llama/Llama-3.2-1B-Instruct checkpoint [1], with its core weights kept frozen throughout all stages. We use the LLaMA-Factory [12] framework and the model's original tokenizer without modification. Training is conducted in bf16 precision, accelerated by FlashAttention, and leverages the native RoPE positional encoding. The context length is set to 8192 for CPT and 4096 for SFT and DPO. All modifications are performed exclusively via LoRA adapters, which we apply to all linear layers in the attention and MLP blocks.

*4.3 Data & Preprocessing*

*4.3.1. CPT Corpus*

Our CPT data compose of two publicly available Traditional Chinese datasets, chosen to provide both broad linguistic coverage and specific domain knowledge. The primary component is the **Taiwan-Text-Excellence-2B dataset** [13], a large-scale collection of high-quality news and articles. This general-domain corpus consists of approximately 1.78 million documents and contains roughly 2 billion tokens, offering extensive exposure to common prose. To enhance the model's capabilities in a specialized domain, we supplemented this with **the republic_of_china_judgements_4_continue_pretrain dataset** [14], a ~300 MB collection of Taiwanese legal judgments.

According to publishers, both source datasets were pre-processed, include document-level deduplication, length-filtering and general text cleaning. Therefore, we did not apply any additional script normalization, or any code-switch filtering.

*4.3.2. SFT Dataset*

Our SFT dataset is a synthetic collection of 1,500 examples, generated via GPT-4-mini using curated prompts. It comprises five distinct instruction-following tasks, with 300 examples uniformly sampled for each. All data entries follow an (instruction, input, output) schema and are formatted using the Llama 3 instruction template. The design of these tasks is guided by three core principles: (i) enforcing single-language output (no code-switching), (ii) ensuring faithfulness to the provided context, and (iii) controlling the structure of long-form generation.

The five tasks are as follows:

1. **Chinese to English Translation** - enforces one-to-one semantic mapping to calibrate a "no add/no drop" prior
2. **English to Chinese Translation -** Trains the model to generate pure Traditional Chinese, preventing code-switching and ensuring correct punctuation.
3. **Faulty-Premise Question Answering** - Requires the model to first identify and correct a flawed premise in a question, then answer it using only information from a given passage.
4. **Grounded Self-Ask/Answer** - Prompts the model to generate question-answer pairs based solely on a provided text, reinforcing evidence-based reasoning.
5. **Hierarchical NER and Outline Extraction** - Tests the model's ability to produce structured, long-form output by requiring it to perform named-entity recognition (NER) and organize the results into a hierarchical outline.

*4.3.3. DPO dataset*

Our DPO dataset is a pairwise preference corpus derived from the five instruction tasks used in the SFT stage. For each prompt $x_i$ (instruction + input), we generate two contrasting responses.

The preferred response ($y_i^+$) is generated by GPT-4-mini, curated to be linguistically pure (Traditional Chinese only) and semantically faithful. The rejected response ($y_i^-$) is an output from our own CPT+SFT model, specifically chosen for exhibiting the artifacts we aim to eliminate, such as code-switching or faithfulness violations.

This process yields a dataset of preference triplets, formally defined as [3]:

$$D_{DPO} = \{(x_i, y_i^+, y_i^-)\}_{i=1}^{N}$$

*4.4 Training Recipe*

All training stages use the AdamW optimizer with a cosine learning rate schedule and a 3% warmup period. We use a per-device batch size of one with gradient accumulation.

*4.4.1. Continual Pre-training (CPT)*

- **Objective**: To shift the model's language priors toward Traditional Chinese (TC) using a causal language modeling objective on a frozen base model.

- **Setup**: This stage consists of two steps. **Step 1 (general corpus)** uses a learning rate of 6e-6. **Step 2** (specialized corpus) uses a learning rate of 1e-5. Both use a context length of 8192, LoRA rank r=512, α=1024, dropout of 0.4, and enable packing.

- **Rationale**: A high-rank LoRA provides the capacity needed for language adaptation without unfreezing the base model. The long context length prepares the model for long-prompt tasks in later stages.

*4.4.2. Supervised Finetuning (SFT)*

- **Objective**. To align the model with specific instruction formats and improve its ability to generate contextually faithful, structured outputs.

- **Setup**. This stage continues training the high-rank adapter from CPT. The context length is 4096, LoRA rank r=512, α=1024, dropout is 0.4, and the learning rate is 7e-7. Packing is disabled.

- **Rationale**. The high-rank adapter preserves the capacity for structural and long-context behaviors learned in CPT. Disabling packing is crucial for instruction tuning to avoid cross-sample interference.

*4.4.3. Direct Preference Optimization*

- **Objective**. To refine the model by optimizing for a preference for linguistically pure TC outputs over semantically equivalent answers that contain code-switching.

- **Setup**. The SFT adapter is merged into the base model. A new, low-rank ("thin") LoRA adapter is then trained with rank r=12, α=24, dropout of 0.4, and a learning rate of 6e-6.

- **Rationale**. Freezing the SFT-merged base protects the capabilities learned during instruction tuning. A low-rank DPO adapter provides efficient preference steering without risking catastrophic forgetting [3].

# 5. Evaluation Protocol

This section outlines the methodology for evaluating model performance. We define the scope of our investigation, the fixed configuration for text generation, and the precise policy for classifying language purity.

*5.1. Scope & Hypotheses*

The central goal of our evaluation is to determine if the fine-tuned model produces text that is exclusively Traditional-Chinese (TC), demonstrating minimal contamination and stable linguistic purity. To this end, we formulate two primary hypotheses:

- (H1): the fine-tuned model yields a lower Other-Language Rate (OLR) than the base model.
- (H2) the fine-tuned model will achieve a significantly higher Pass@TC rate, indicate a perfect pure outputs.

*5.2. Inference Configuration*

To ensure a fair and reproducible comparison, all text generation for both the base and fine-tuned models was conducted under a single, fixed decoding configuration. The parameters were:

- Temperature 0.2
- Top-p: 0.9
- Max New Tokens: 1024
- Repetition Penalty: 1.05.

These conservative settings were chosen to minimize randomness and favor high-fidelity, deterministic outputs, making the evaluation a stricter test of the models' learned knowledge rather than their creative variance.

*5.3. Language Policy (TC-only)*

To objectively measure contamination, we established a strict, character-level language policy. A character is considered TC-legal if it is either:

1. **Han ideograph** (CJK Unified Ideographs + Extensions, plus Compatibility Ideographs)
2. **Decimal Digit**, as defined by the Unicode property General Category=Nd.

This excludes Latin letters, most punctuation, and special symbols, setting **a high bar for linguistic purity.**

Before scoring, each generated output was normalized via:

1. **Unicode Normalization**: NFKC normalization is applied to standardize character forms. A key effect is the conversion of full-width digits (e.g., 5) to their ASCII equivalents (e.g., 5), which are then accepted as legal characters.
2. **Whitespace Unification**: All whitespace sequences are collapsed into a single space.
3. **No Lowercasing**: Case is preserved to ensure that any Latin letters are caught as non-compliant characters.

*5.4. Metrics*

To quantitatively assess the model's adherence to the Traditional-Chinese (TC) language policy (§5.3) under the fixed decoding settings (§5.2), we define two primary metrics: one to measure the degree of contamination and another to measure the frequency of perfect outputs.

*5.4.1. Other-Language Rate (OLR)*

This metric quantifies the density of contamination in an output. It is calculated as the fraction of characters in the generated string that are not compliant with the TC-only policy. The formula is:

$$\text{OLR}_{\text{char}}(S) = \frac{\sum_{c \in S} \mathbf{o1}[\neg \text{is\_TC}(c)]}{|S|_{\text{chars}}}$$

*5.4.2. Pass@TC (Higher is better)*

Measures the frequency of perfectly clean outputs:

$$\text{Pass@TC}(S) = \mathbb{1}[\text{OLR}(S) = 0]$$

This indicator function returns 1 for a perfectly clean output and 0 otherwise. This metric directly evaluates the model's reliability in generating 100% pure TC text, providing a clear success rate for our primary objective.

# 6. Experiments

This section details the experimental setup, including the models, datasets used for both training and evaluation, and the precise procedure followed to obtain our results.

*6.1. Experimental Setup*

We benchmarked four models under identical decoding conditions (§5.2):
- Llama-3.2-1B-Instruct (Base) – zero-shot baseline.

- PureTC-1B (Ours) – after CPT → SFT → DPO pipeline.
- Llama-3B – a larger open-weight model of the same family.
- Qwen-1.5B – a strong open-weight baseline with similar scale.

The evaluation was conducted on a prompt pool of approximately 605 items generated by GPT-4o-mini, categorized into eight task families. Each family is crafted to stress a distinct capability relevant to TC stability:

1. **EN->TC Translation:** Translating English sentences contain names/acronyms to probe literal fidelity.
2. **Bilingual Purge:** Rewrite a mixed Chinese/English paragraph into pure TC to test code-switch removal.
3. **SC->TC Conversion:** Simplified-Chinese paragraph → Traditional to probe clean character-set conversion without leakage.
4. **Structured JSON:** Creating structured JSON with Chinese keys/values to test structural generation under purity constraints.
5. **Content Organization:** Producing bulleted lists or outlines to test content organization.
6. **Long-form Summary:** Summarizing a long Chinese text to probe stability over extended contexts.
7. **Noise Robustness:** Responding to Chinese text mixed with non-linguistic tokens (URLs, code, emojis) to test the model's ability to ignore noise.
8. **Entity Translation:** Translating proper-noun-dense English text to probe entity rendering consistency.

*6.2. Evaluation Procedure*

For each of the ≈605 prompts, the following steps were applied:

1. **Generation** – Each model produced outputs under the fixed decoding setup (§5.2).
2. **Normalization** – Outputs were standardized with Unicode NFKC normalization and whitespace unification (§5.3)
3. **Scoring** – Outputs were evaluated with **OLR** (contamination density) and **Pass@TC** (strict purity).

Macro- vs. Micro-Averaging

Because prompts are not uniformly distributed across task families (e.g., 100 for Bilingual Purge vs. 60 for Long-form Summary), we report two levels of aggregation:

Macro-average: Arithmetic mean across the eight families, giving each task equal weight.

- Micro-average: Weighted mean across all 605 prompts, reflecting overall token-level performance.
- This dual reporting captures both **family-level robustness** and **global reliability**, ensuring that improvements are not biased by task imbalance.

## 7. Result

### 7.1. Headline Findings

Across ≈605 prompts, PureTC-1B consistently outperforms the base model and rivals larger models:

- **Macro-average (per-family unweighted)**: OLR reduced from 0.214 → 0.098 (−66.8% rel.), Pass@TC improved from 9.5% → 29.9% (+20.4 pp).
- **Micro-average (weighted by family size)**: OLR reduced from 0.231 → 0.113 (−51.3% rel.), Pass@TC improved from 10.4% → 30.3% (+19.9 pp).

These gains are robust across 7 of 8 task families, with the sole regression observed on the Noise Robustness task.

### 7.2. Detailed Performance Analysis

A detailed A detailed breakdown by task family is shown in Table 1, comparing the base and fine-tuned 1B model. The results show three clear patterns:

- **On Named Entity Translation,** PureTC-1B reduces non-TC contamination by 77.2% vs. Llama-3B and 57.2% vs. Qwen-1.5B.
- **On Bilingual Purity** and **SC→TC Conversion,** PureTC-1B achieves higher strict-purity rates than both larger models, despite its smaller scale.
- Only on **Noise Robustness** does Llama-3B outperform, as PureTC-1B tends to over-translate noise rather than ignore it.

| Task | N | **Olr PassTC-1b** | **Pass PassTC-1B** | Olr Llama-1b | **Pass Llama-1b** | Olr Llama-3b | **Pass Llama-3b** | Olr Qwen-1.5b | **Pass Qwen-1.5b** |
|---|---|---|---|---|---|---|---|---|---|

13 of 17| | | | | | | | | |
|---|---|---|---|---|---|---|---|---|
| Entity Translation | 80 | **0.123** | **12.0%** | 0.310 | 1.3% | 0.243 | 4.0% | 0.185 | 10.7% |
| Bilingual Purity | 100 | **0.016** | **59.6%** | 0.125 | 33.3% | 0.131 | 37.4% | 0.071 | 55.6% |
| Sc Conversion | 60 | **0.014** | **46.3%** | 0.062 | 14.8% | 0.232 | 3.7% | 0.040 | 57.4% |
| Structured Json | 100 | **0.078** | **46.2%** | 0.302 | 15.4% | 0.476 | 10.6% | 0.061 | 19.2% |
| Content Organization | 60 | **0.014** | **20.0%** | 0.123 | 0.0% | 0.151 | 4.4% | 0.043 | 37.8% |
| Long-Form Summary | 60 | **0.017** | **26.7%** | 0.130 | 2.2% | 0.170 | 2.2% | 0.067 | 40.0% |
| Noise Robustness | 100 | **0.467** | **0.0%** | 0.389 | 5.6% | 0.329 | 23.6% | 0.480 | 0.0% |
| Named Entity Translation | 100 | **0.057** | **28.7%** | 0.273 | 3.2% | 0.250 | 6.4% | 0.133 | 12.8% |
| Macro-Avg | — | **0.098** | **29.9%** | 0.214 | 9.5% | 0.248 | 11.9% | 0.135 | 29.2% |
| Micro-Avg | 660 | **0.104** | **30.3%** | 0.231 | 10.4% | 0.257 | 11.8% | 0.145 | 31.3% |

**Table 1:** Comprehensive Evaluation Results

## 7.3. Baseline Comparison

Building on this, **Table 2** compares PureTC-1B against **larger baselines** (Llama-3B, Qwen-1.5B). The data highlights that stability is **not simply a function of scale**:

- PureTC-1B outperforms Llama-3B and Qwen-1.5B on most families, especially Named Entity Translation (77.2% relative reduction vs. Llama-3B, 57.2% vs. Qwen-1.5B).
- Larger models still perform better on Noise Robustness, indicating that size confers some generalization ability not fully captured by TC-specialization.

| TASK | OLR VS. 1B (%↓) | PASS VS. 1B (PP) | OLR VS. 3B (%↓) | PASS VS. 3B (PP) | OLR VS. QWEN 1.5B (%↓) | PASS VS. QWEN 1.5B (PP) |
|---|---|---|---|---|---|---|
| Entity Translation | +60.4% | +10.7 pp | +49.4% | +8.0 pp | +33.6% | +1.3 pp |
| Bilingual Purity | +87.2% | +26.3 pp | +87.8% | +22.2 pp | +77.4% | +4.0 pp |
| Sc Conversion | +78.3% | +31.5 pp | +94.0% | +42.6 pp | +65.0% | -11.1 pp |
| Structured Json | +74.1% | +30.8 pp | +83.6% | +35.6 pp | -27.9% | +27.0 pp |
| Content Organization | +88.3% | +20.0 pp | +90.7% | +15.6 pp | +67.5% | -17.8 pp |
| Long-Form Summary | +86.5% | +24.4 pp | +90.0% | +24.5 pp | +74.5% | -13.3 pp |
| Noise Robustness | -20.1% | -5.6 pp | -41.9% | -23.6 pp | +2.7% | +0.0 pp |
| Named Entity Translation | +79.3% | +25.5 pp | +77.2% | +22.3 pp | +57.2% | +16.0 pp |
| Macro-Avg | **+66.8%** | **+20.4 pp** | **+69.5%** | **+18.9 pp** | **+44.7%** | **+5.3 pp** |
| Micro-Avg | **+51.3%** | **+19.9 pp** | **+59.6%** | **+18.5 pp** | **+28.4%** | **-1.0 pp** |

Table 2: Comparative Performance of Fine-tuned vs. Baseline Models

### 7.4. Summary of Results

Collectively, the results confirm that PureTC-1B substantially improves TC stability at the 1B scale, reducing contamination by over 50% (micro-avg OLR) and tripling the strict-purity rate (Pass@TC). These gains are robust across 7 of 8 task families, with the only regression observed in noise handling. Importantly, PureTC-1B rivals or outperforms larger models (Llama-3B, Qwen-1.5B) on most stability-oriented tasks, demonstrating that stability is not solely a function of scale but can be systematically enforced through our CPT→SFT→DPO pipeline.

## 8. Discussion & Conclusion

### 8.1 Discussion

Our experiments demonstrate that **PureTC-1B** achieves substantial gains in Traditional Chinese (TC) token stability. However, several limitations and broader implications deserve discussion.

**Limitations.**

First, while our model significantly reduces non-TC contamination, **Noise Robustness** tasks revealed a regression: the fine-tuned system tends to over-translate or transliterate noisy inputs (URLs, code, emojis) instead of ignoring them. Second, although our evaluation is framed around **minimizing Other-Language Rate (OLR),** we note that a **small amount of English tokens can be reasonable and even necessary** (e.g., proper nouns, technical acronyms). Thus, the goal is not to eliminate English entirely but to **control leakage** so that OLR remains within an acceptable range for professional applications. Third, the evaluation dataset is imbalanced across task families, which required reporting both macro- and micro-averages. Finally, while we benchmarked against Llama-3B and Qwen-1.5B, larger-scale baselines (e.g., 7B+ models) remain unexplored due to hardware constraints.

**Applicability.**

Despite these limitations, PureTC-1B is **adapter-only, hardware-friendly, and reproducible**, making it deployable on 12 GB GPUs and thus accessible to small labs, public institutions, and startups. This practicality sets it apart from larger models, which are often inaccessible for real-world deployments. Moreover, our methodology—CPT to strengthen language priors, SFT for task alignment, and DPO for preference correction—is not inherently TC-specific. It could generalize to other non-English languages facing similar token instability, offering a **blueprint for language stabilization in resource-constrained settings.**

*8.2 Conclusion*

In this work, we introduced **PureTC-1B**, a three-stage stabilization pipeline for SLMs that enforces Traditional Chinese adherence without full-model retraining. Our approach systematically combines **Continual Pre-Training (CPT), Supervised Fine-Tuning (SFT),** and **Direct Preference Optimization (DPO)** under LoRA adapters.

Our contributions are threefold:

1. We formalized **token-level instability** in TC generation as a measurable deployment risk and proposed new evaluation metrics **(OLR, Pass@TC).**
2. We designed a **modular adapter-based pipeline** that is reproducible, hardware-friendly, and effective at stabilizing SLMs.
3. We demonstrated that PureTC-1B achieves **over 50% relative reduction in contamination (micro-average OLR)** and nearly **triples strict-purity success (Pass@TC)**, outperforming even larger baselines (Llama-3B, Qwen-1.5B) on most tasks.

Looking forward, two directions stand out:

- **Dataset refinement:** constructing balanced and more diverse evaluation sets, including multilingual noise and adversarial prompts.
- **Scalability:** applying the stabilization pipeline to larger base models and testing whether stability gains scale proportionally.

In summary, this thesis demonstrates that **linguistic stability can be systematically engineered in small models** without sacrificing deplorability. PureTC-1B is not only a step toward reliable TC applications, but also a case study in how **small, specialized LMs** can meet real-world requirements where scale alone does not suffice.